\documentclass[a4paper, 10pt, conference]{ieeeconf} 

\IEEEoverridecommandlockouts    

\usepackage{graphics} 
\usepackage{epsfig} 
\usepackage{mathptmx} 
\usepackage{times} 
\usepackage{amsmath} 
\usepackage{amssymb} 
\usepackage{color}
\usepackage{xcolor}
\usepackage{booktabs,siunitx} 
\usepackage{svg}
\usepackage{multirow}
\usepackage{rotating}
\usepackage[export]{adjustbox}
\usepackage{array, makecell}
\usepackage{hyperref}
\usepackage[normalem]{ulem}

\usepackage{graphicx}
\usepackage{support-caption}
\usepackage{subcaption}
\usepackage{mwe}

\usepackage{bm}
\usepackage{balance}

\usepackage{nicefrac,xfrac}

\usepackage{pifont}
\newcommand{\cmark}{\ding{51}}%
\newcommand{\xmark}{\ding{55}}%

\newcommand{\eg}{\textit{e}.\textit{g}. }
\newcommand{\etal}{\textit{et al. }}

\useunder{\uline}{\ul}{}

\usepackage[symbol]{footmisc}

\title{\LARGE \bf
Quantifying Data Augmentation for LiDAR-based 3D Object Detection 
}

\author{Martin Hahner$^{1}$, Dengxin Dai$^{1}$, Alexander Liniger$^{1}$, and Luc Van Gool$^{1,2}$
\thanks{$^{1}$Martin Hahner, Dengxin Dai, Alexander Liniger and Luc Van Gool are all with the Toyota TRACE-Zurich team  at the Computer Vision Lab,  ETH~Zurich, 8092 Zurich, Switzerland. 
 {\tt\small firstname.lastname@vision.ee.ethz.ch}}%
\thanks{$^{2}$Luc Van Gool is also with the Toyota TRACE-Leuven team at the Dept. of Electrical Engineering ESAT, KU Leuven, 3001 Leuven, Belgium.
 {\tt\small luc.vangool@kuleuven.be}}%
}

\begin{document}

\maketitle
\thispagestyle{empty}
\pagestyle{empty}

\begin{abstract}

In this work, we shed light on different data augmentation techniques commonly used in Light Detection and Ranging (LiDAR) based 3D Object Detection. For the bulk of our experiments, we utilize the well known \textit{PointPillars}~\cite{PointPillars} pipeline and the well established KITTI~\cite{KITTI} dataset. We investigate a variety of \textit{global} and \textit{local} augmentation techniques, where \textit{global} augmentation techniques are applied to the entire point cloud of a scene and \textit{local} augmentation techniques are only applied to points belonging to individual objects in the scene. 
Our findings show that both types of data augmentation can lead to performance increases, but it also turns out, that some augmentation techniques, such as individual object translation, for example, can be counterproductive and can hurt the overall performance. 
We show that these findings transfer and generalize well to other state of the art 3D Object Detection methods and the challenging STF~\cite{STF} dataset.
On the KITTI~\cite{KITTI} dataset we can gain up to $\textbf{1.5\%}$ and on the STF~\cite{STF} dataset up to $\textbf{1.7\%}$ in 3D mAP$_{\textbf{40}}$ on the moderate car class. 

\end{abstract}

\section{Introduction} \label{introduction}

The last years have seen tremendous progress in 3D Object Detection for autonomous driving and the task really only emerged in 2017, when the KITTI~\cite{KITTI} dataset, originally introduced in 2012, was extended by novel benchmarks for 3D Object Detection including 3D and bird's eye view (BEV) evaluation. Since then, many more publicly available LiDAR datasets with 3D bounding box annotations have followed. nuScenes~\cite{nuScenes}, the Lyft~Level5~Perception~Dataset~\cite{LyftL5Dataset} and STF~\cite{STF} are just some of them, others include \cite{Apollo3D, Argoverse, Honda3D} and \cite{WaymoOpenDataset}. With LiDAR sensors getting cheaper \cite{LiDAR} and thereby becoming a viable option for autonomous driving, these LiDAR datasets are fundamental to improve the current state of the art in LiDAR based 3D Object Detection.

Further, the motivation behind the task of 3D Object Detection is that autonomous driving cars need to find a trajectory in the real world (a 3D space). Hence, 2D Object Detection is insufficient since 2D Object Detection only delivers a location and dimensions in the image plane. Even if 2D Object Detection results are combined with Monocular Depth Estimation methods such as \cite{packnet}, the performance is not as high as directly reasoning in 3D LiDAR point clouds. Also, Monocular 3D Object Detection methods like Pseudo-LiDAR~\cite{Pseudo-LiDAR} often re-purpose 3D Object Detection pipelines meant for point cloud processing. This indicates that point clouds are indeed a much better representation to detect objects in 3D compared to utilizing purely image based pipelines. The authors of Pseudo-LiDAR~\cite{Pseudo-LiDAR} propose to convert image-based depth maps into pseudo-LiDAR representations (essentially fake point clouds). Newer state of the art Monocular 3D Object Detection methods that are not relying on such pseudo-LiDAR point clouds anymore, like DD3D~\cite{dd3d} still lack far behind in performance compared to state of the art LiDAR based object detectors, \eg PV-RCNN~\cite{PV-RCNN} (16.34\% vs. 81.43\% 3D mAP on moderate cars in the KITTI~\cite{KITTI} dataset, respectively). Image based 3D Object Detection might just be too ill-posed, which makes us believe that in order to reach the desired level of safety for autonomous driving cars, we need a sensor on-board that is able to directly measure raw 3D depth information.

To specify the task at hand, what we are looking for in 3D Object Detection for autonomous driving are specifically seven degrees of freedom as opposed to only four in 2D Object Detection. In the task of 3D Object Detection we want to predict the center position of an object $x_c$, $y_c$, $z_c$, its dimensions $w$, $l$, $h$ and the \textit{yaw} angle $\theta$ (rotation around the upright axis). Since data augmentation for 3D Object Detection has been found crucial by many works~\cite{PIXOR, VoxelNet, SECOND}, 
we want to investigate data augmentation for LiDAR based 3D Object Detection in great detail. Given our findings (presented in Section~\ref{experiments}), we believe that such an extensive augmentation study has been long overdue. 

The contributions in our paper are three-fold, first and foremost, we present an in-depth study of augmentation methods for LiDAR based 3D Object Detection. This study allows practitioners to short cut time consuming experiments and get good results quicker. Second, we show some non-intuitive results, for example, translation of objects can reduce performance, or that excluding \textit{hard} cases from training can increase the performance for these \textit{hard} cases during evaluation. These findings pose new research questions and suggest some insight how 3D Object Detection networks might work internally. Finally, based on our study of augmentation methods, we propose a new augmentation policy that is able to increase the performance of several state of the art 3D Object Detection methods on two different LiDAR datasets: KITTI~\cite{KITTI} and STF~\cite{STF}. 

\section{Related Work} \label{related_work}

In this Section, we review relevant work in the field of LiDAR-based 3D Object Detection (Section \ref{detection}) and LiDAR Data Augmentation (Section \ref{augmentation}).

\subsection{LiDAR-based 3D Object Detection} \label{detection} 

The earliest works in this line of work include MV3D~\cite{MV3D}, PIXOR~\cite{PIXOR} and AVOD~\cite{AVOD}. They all divide the 3D space into a voxel grid. Then they use hand-crafted features (\eg point count per voxel) and hand-crafted neighborhood features (\eg maximum point count per pillar). In the end, they have a 3D tensor of neighborhood features $(x, y, c)$ and apply regular 2D convolutions on them. The drawback of these methods mostly lies in the hand-crafted feature design.

PointNet~\cite{PointNet} laid the foundations for many applications, not just for 3D Object Detection. Their work can also be applied to point cloud Classification, Part Segmentation and Semantic Segmentation. The main reason their work is so applicable to multiple domains is because its architecture is relatively lightweight, yet highly efficient and effective, it can process up to one million points per second.

VoxelNet~\cite{VoxelNet} is one of the seminal works in the area of 3D Object Detection, as a first, this work presents a 3D Object Detection pipeline in an end-to-end fashion without any hand-crafted features. They apply a PointNet-like~\cite{PointNet} architecture to every individual voxel leading to a 4D tensor $(x, y, z, c)$. Then they perform computationally intensive 3D convolutions to consolidate the $z$-dimension, yielding a 3D tensor $(x, y, c')$. After this consolidation step, they only use computationally much cheaper 2D convolutions to generate and regress their region proposals.

Another significant contribution to the area of LiDAR based 3D Object Detection is SECOND~\cite{SECOND}, they were able speed up the bottleneck of earlier works: 3D convolutions, by proposing sparse 3D convolutions.
Building on top of~\cite{SECOND}, PointPillars~\cite{PointPillars} gets entirely rid of the $z$-dimension by dividing the 3D space into pillars instead of voxels. This change by itself gives a 10-100x speed-up compared to VoxelNet~\cite{VoxelNet}. Additionally, to further speed up the novel encoder, they process each pillar only using a single $1x1$ convolution + max-pooling layer instead of a more complex PointNet-like~\cite{PointNet} architecture.

Shi \etal achieved several recent milestones in 3D Object Detection. PointRCNN~\cite{PointRCNN} is a two-stage architecture, where the first stage generates 3D bounding box proposals from a point cloud in a bottom-up manner and the second stage refines these 3D bounding box proposals in a canonical fashion. Part-A$^2$~\cite{PartA2} is part-aware in a sense that the network takes into account which part of the object a point belongs to. It leverages these intra-object part locations and can thereby achieve better results. PV-RCNN~\cite{PV-RCNN} and it's successor PV-RCNN++~\cite{shi2021pvrcnn} are the latest of their works that simultaneously process (coarse) voxels and the raw points of the point cloud at the same time.

CenterPoint~\cite{CP} is a two-stage architecture, where the first stage predicts centers of objects using a key-point detector and the second stage thereafter regresses 3D size and orientation of these objects.
VoteNet~\cite{VoteNet} has not proven itself yet on an automotive dataset, but showed promising
results on two indoor scene understanding datasets, Scan-Net~\cite{ScanNet} and SUN RGB-D~\cite{SUN}.

\subsection{LiDAR Data Augmentation} \label{augmentation} 

Ever since SECOND~\cite{SECOND} introduced ground truth (GT) sampling (a.k.a. \textit{GT-Sampling}, see Section~\ref{GT-Sampling}), data augmentation has been widely adopted for 3D Object Detection. PA-AUG~\cite{Part-Aware} extends \textit{GT-Sampling} by dividing the GT objects into sub-partitions and randomly augmenting each sub-partition, \eg dropping all points in a partition. They demonstrate that their augmentation technique can improve the performance of PV-RCNN~\cite{PV-RCNN} by 0.47\% on moderate cars in the KITTI~\cite{KITTI} validation split.
Similarly,~\cite{zheng2021se} divide each ground truth object into six pyramids, one inward facing pyramid for each face of the 3D bounding box, and then independently augments each pyramid
by random dropout, swap, and/or sparsify operations. This way, they can successfully boost the performance of their SE-SSD detector by 0.31\% on moderate cars in the KITTI~\cite{KITTI} validation split.
\cite{Pattern-Aware} also builds on top of \textit{GT-Sampling}, they propose an algorithm to mimic the diverging point pattern that occurs when objects are placed at different distances (due to diverging LiDAR beams) and can improve the performance of PV-RCNN~\cite{PV-RCNN} by 0.29\% on moderate cars in the KITTI~\cite{KITTI} validation split. 

PointPainting~\cite{PointPainting} and PointAugmenting~\cite{PointAugmenting} are another line of work that can also be considered as LiDAR data augmentation. The idea of PointPainting~\cite{PointPainting} is to run an Image Segmentation network on the camera image first and then project the predicted segmentation class scores onto the LiDAR point cloud. PointAugmenting~\cite{PointAugmenting} takes this idea one step further and projects deep features of a 2D Object Detector onto the LiDAR point cloud instead. In both cases, the resulting augmented point cloud can still be processed by standard 3D Object Detection methods introduced in Section~\ref{detection}, resulting in a higher performance compared to processing the original, non-augmented LiDAR point clouds.

Some works use physical modelling to augment clear weather point clouds with artificial fog~\cite{HahnerICCV21} or artificial snowfall~\cite{HahnerCVPR22} and show improvements on point clouds collected in such real-world adverse conditions. 

Also other fields that process point clouds, like point cloud classification make use of augmentation. PointMixup~\cite{PointMixup} \eg interpolates point clouds from different classes to regularize the training process and 
PointAugment~\cite{PointAugment} provides an auto-augmentation framework for this task.

While in this paper, we investigate manually designed standard data augmentation policies, PPBA~\cite{PPBA} attempts to automate this design process via an evolutionary algorithm. Their search algorithm narrows down the search space every iteration and adopts the best parameters discovered in previous iterations.

\begin{figure*}
    \centering
    \begin{subfigure}[b]{0.475\textwidth}
        \centering
        \includegraphics[width=\textwidth]{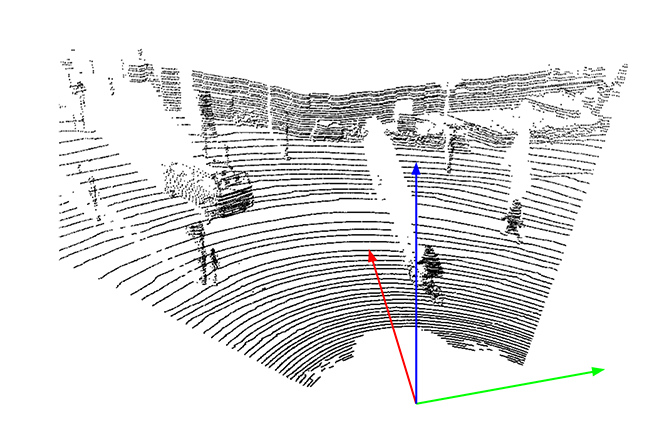}
        \caption[]%
        {{\small Original scene.}}    
        \label{fig:original_scene}
    \end{subfigure}
    \hfill
    \begin{subfigure}[b]{0.475\textwidth}  
        \centering 
        \includegraphics[width=\textwidth]{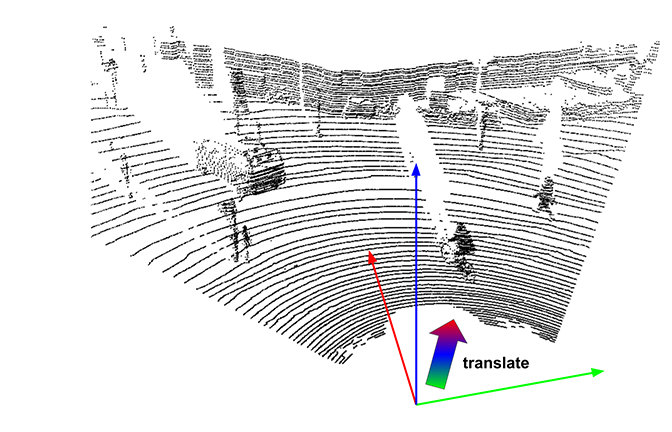}
        \caption[]%
        {{\small \textit{Global Translation}~(section~\ref{global_translation}).}}    
        \label{fig:global_translation}
    \end{subfigure}
    \vskip\baselineskip
    \begin{subfigure}[b]{0.475\textwidth}   
        \centering 
        \includegraphics[width=\textwidth]{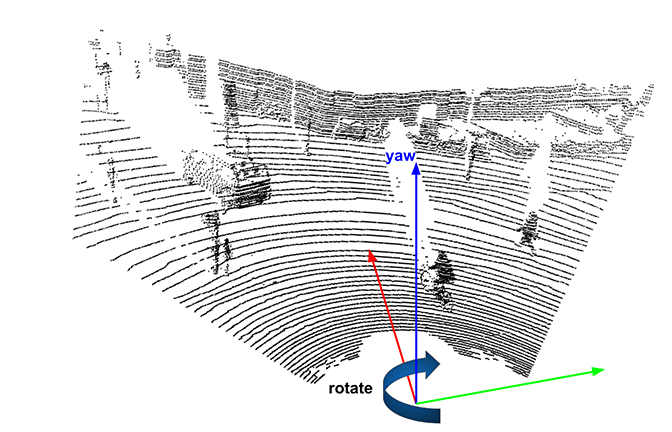}
        \caption[]%
        {{\small \textit{Global Rotation}~(section~\ref{global_rotation}).}}    
        \label{fig:global_rotation}
    \end{subfigure}
    \quad
    \begin{subfigure}[b]{0.475\textwidth}   
        \centering 
        \includegraphics[width=\textwidth]{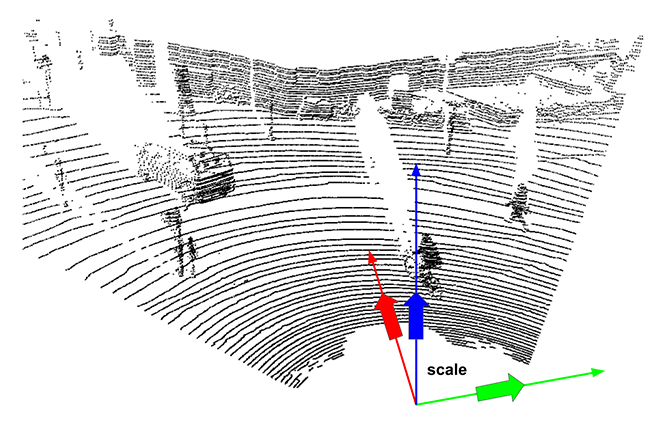}
        \caption[]%
        {{\small \textit{Global Scaling}~(section~\ref{global_scaling}).}}    
        \label{fig:global_scaling}
    \end{subfigure}
    \vskip\baselineskip
    \begin{subfigure}[b]{0.475\textwidth}   
        \centering 
        \includegraphics[width=\textwidth]{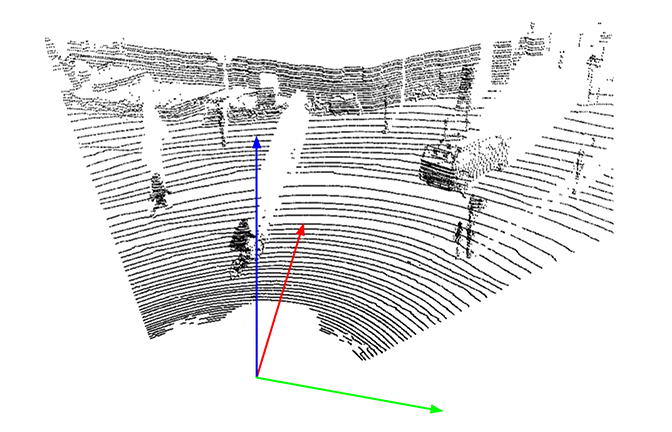}
        \caption[]%
        {{\small \textit{Random Flip}~(section~\ref{random_flip}).}}    
        \label{fig:random_flip}
    \end{subfigure}
    \quad
    \begin{subfigure}[b]{0.475\textwidth}   
        \centering 
        \includegraphics[width=\textwidth]{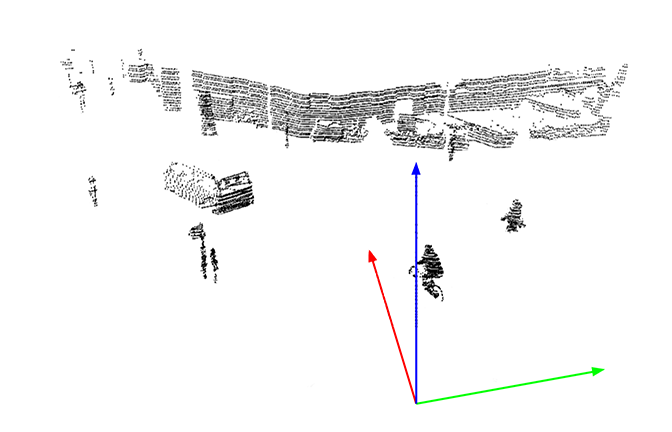}
        \caption[]%
        {{\small \textit{Ground Removal}~(section~\ref{ground_removal}).}}    
        \label{fig:ground_removal}
    \end{subfigure}
    \caption[]%
    {Visualization of all \textit{global} augmentation techniques investigated in this paper.} 
    \label{fig:global_augmentation}
\end{figure*}
\begin{figure*}
    \centering
    \begin{subfigure}[b]{0.225\textwidth}
        \centering
        \includegraphics[width=\textwidth]{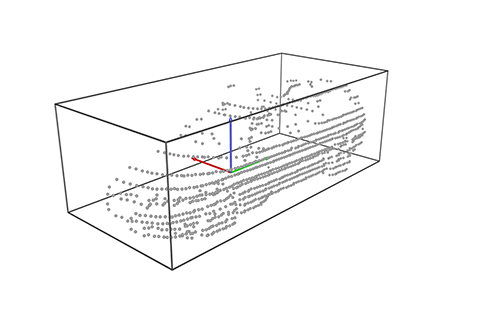}
        \caption[]%
        {{\small Original annotation. \\ \color{white}ghost}}    
        \label{fig:original_annotation}
    \end{subfigure}
    \hfill
    \begin{subfigure}[b]{0.225\textwidth}  
        \centering 
        \includegraphics[width=\textwidth]{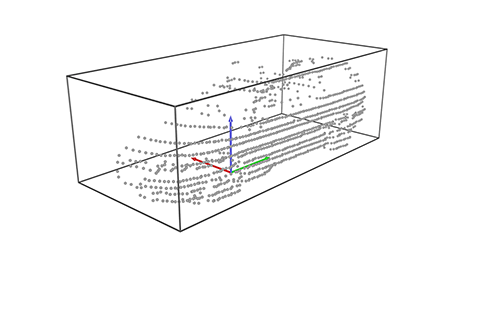}
        \caption[]%
        {{\small \textit{Local Translation} \\ (section~\ref{local_translation}).}}    
        \label{fig:local_translation}
    \end{subfigure}
    \hfill
    \begin{subfigure}[b]{0.225\textwidth}  
        \centering 
        \includegraphics[width=\textwidth]{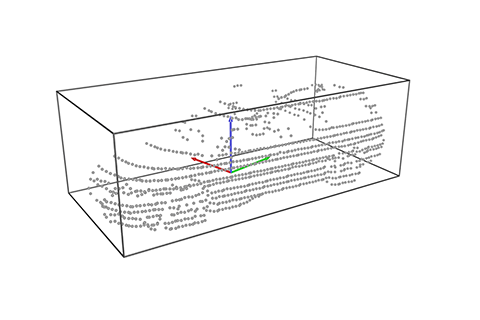}
        \caption[]%
        {{\small \textit{Local Rotation} \\ (section~\ref{local_rotation}).}}    
        \label{fig:local_rotation}
    \end{subfigure}
    \hfill
    \begin{subfigure}[b]{0.225\textwidth}  
        \centering 
        \includegraphics[width=\textwidth]{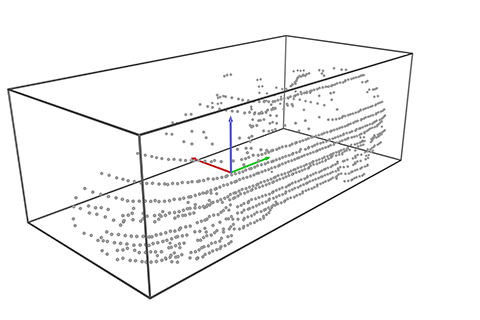}
        \caption[]%
        {{\small \textit{Local Scaling} \\ (section~\ref{local_scaling}).}}    
        \label{fig:local_scaling}
    \end{subfigure}
    \caption[]%
    {Visualization of all \textit{local} augmentation techniques investigated in this paper.} 
    \label{fig:local_augmentation}
\end{figure*}

\section{Method} \label{method}
In this Section, we present all augmentation techniques investigated in this paper. They can be categorized into four categories: \\
- \textit{global} augmentations (Section \ref{global_augmentation}),  \\
- \textit{local} augmentations (Section \ref{local_augmentation}), \\
- \textit{GT-Filtering} (Section \ref{filter_augmentations}) and \\
- \textit{GT-Sampling} (Section \ref{GT-Sampling}).

\subsection{Global Augmentations} \label{global_augmentation}  
\textit{Global} augmentations are applied to the all points in the point cloud $P = \{p_0, ..., p_n\}$ and all annotations $A = \{a_0, ..., a_m\}$ simultaneously. A visualization of all \textit{global} augmentation techniques investigated in this paper are shown in Fig.~\ref{fig:global_augmentation} applied to an example scene~(Fig.~\ref{fig:original_scene}). \\

\vspace{-3mm}
\subsubsection{Global Translation} \label{global_translation} \hfill\\
Global translation~(Fig.~\ref{fig:global_translation}) means that we are translating every point $p(x, y, z) \in P$ such that an augmented point $p^*$ has the form $p(x + \Delta x, y + \Delta y, z + \Delta z)$. Simultaneously, we shift every annotation $a(x_c, y_c, z_c, w, l, h, \theta) \in A$ such that an augmented annotation $a^*$ has the form $a(x_c + \Delta x, y_c + \Delta y, z_c + \Delta z, w, l, h, \theta)$. Therefore we independently sample $\Delta x, \Delta y$ and $\Delta z$ from a normal distribution $N(0,\sigma^{2})$ where $\sigma$ can take the following values $\sigma^2 \in \{0.1, 0.2, 0.4\}$m. \\

\vspace{-3mm}
\subsubsection{Global Rotation} \label{global_rotation} \hfill\\
Global rotation~(Fig.~\ref{fig:global_rotation}) means that we are rotating every point $p(x, y, z) \in P$ around the upright \textit{yaw} axis by an angle $\alpha$ drawn from a uniform distribution $U(-\beta, + \beta)$ where $\beta \in \{\sfrac{\pi}{8}, \sfrac{\pi}{4}, \sfrac{\pi}{2}\}$. Simultaneously, we rotate every annotation $a(x_c, y_c, z_c, w, l, h, \theta) \in A$ such that the augmented annotation $a^*$ has the form $a(x_c, y_c, z_c, w, l, h, \scriptstyle(\theta + \alpha)\mod 2\pi\textstyle)$. \\

\vspace{-3mm}
\subsubsection{Global Scaling} \label{global_scaling} \hfill\\
Global scaling~(Fig.~\ref{fig:global_scaling}) means that we are scaling every point $p(x, y, z) \in P$ in every direction by a scalar $s$ drawn from a uniform distribution $U(1-t, 1 + t)$ where $t \in \{0.05, 0.1, 0.25\}$ such that an augmented point $p^*$ has the form $p(s \cdot x, s \cdot y, s \cdot z)$. Simultaneously, we scale every annotation $a(x_c, y_c, z_c, w, l, h, \theta) \in A$ such that an augmented annotation $a^*$ has the form $a(s \cdot x_c, s \cdot y_c, s \cdot z_c, s \cdot w, s \cdot l, s \cdot h, \theta)$. \\

\vspace{-3mm}
\subsubsection{Random Flip} \label{random_flip} \hfill\\
Random flip~(Fig.~\ref{fig:random_flip}) means that we flip every point $p(x, y, z) \in P$ by a 50\% chance on the forward facing $x$-axis. So if we flip the whole point cloud, an augmented point $p^*$ has the form $p(x, -y, z)$. Simultaneously, if we flip every point $p(x, y, z) \in P$, we also flip every annotation $a(x_c, y_c, z_c, w, l, h, \theta) \in A$ such that an augmented annotation $a^*$ has the form $a(x_c, - y_c, z_c, w, l, h, \scriptstyle(\theta + \pi)\mod 2\pi\textstyle)$. We do not flip on the sideways facing $y$-axis because KITTI~\cite{KITTI} and STF~\cite{STF} only provide 3D bounding box annotations in the camera field of view. So it does not make sense to flip on the $y$-axis when on the other side, there are no targets to train for. \\

\vspace{-3mm}
\subsubsection{Ground Removal} \label{ground_removal} \hfill\\
Ground removal~(Fig.~\ref{fig:ground_removal}) means that we remove every point $p(x, y, z) \in P$ where $z$ is smaller than a threshold $\epsilon \in$ \{1st, 5th, 10th, 15th\} percentile of all $z$-values in the point cloud $P$. The idea behind this is to get rid of ``background" points, because a LiDAR point cloud is typically very unbalanced in terms of ``foreground" vs. ``background". In the KITTI~\cite{KITTI} training set for example, there are on average 19,047 points per scene in the camera field of view and only 1,382 points ($\approx 7,25\%$) thereof lie inside 3D bounding box annotations (``foreground"). For the KITTI~\cite{KITTI} validation set, this statistic does not look much different, there are on average 18,888 points per scene in the camera field of view and 1,641 points ($\approx 8,69\%$) thereof lie inside 3D bounding box annotations. This augmentation also has the uniqueness, that it is the only augmentation we investigate, that is applied during training and testing. This is to not mess up the distribution of ``foreground" vs. ``background" during the different stages. All other augmentations are only applied during training.

\subsection{Local Augmentations} \label{local_augmentation}
Similar to the \textit{global} augmentation techniques presented in the previous Section, \textit{local} augmentations presented in this Section also involve translation, rotation and scaling. The only difference is, that here we do not apply those transformations to every point $p(x,y,z) \in P$, but only to individual annotations and the points that reside inside those annotations. Thereby we augment every annotation independently from every other, meaning that for every annotation we draw a different random value. A visualization of all \textit{local} augmentation techniques investigated in this paper are shown in Fig.~\ref{fig:local_augmentation} with an example annotation~(Fig.~\ref{fig:original_annotation}). \\

\vspace{-3mm}
\subsubsection{Local Translation} \label{local_translation}\hfill\\
Local translation~(Fig.~\ref{fig:local_translation}) means that we are translating every point $p(x, y, z) \in a$ such that an augmented point $p^*$ has the form $p(x + \Delta x, y + \Delta y, z + \Delta z)$. Simultaneously, we shift the annotation $a(x_c, y_c, z_c, w, l, h, \theta)$ such that an augmented annotation $a^*$ has the form $a(x_c + \Delta x, y_c + \Delta y, z_c + \Delta z, w, l, h, \theta)$. Again, we independently sample $\Delta x, \Delta y$ and $\Delta z$ from a normal distribution $N(0,\sigma^{2})$, with $\sigma^2 \in \{0.05, 0.25, 0.5, 1\}$m. \\

\vspace{-3mm}
\subsubsection{Local Rotation} \label{local_rotation}\hfill\\
Local rotation~(Fig.~\ref{fig:local_rotation}) means that we are rotating every point $p(x, y, z) \in a$ around the upright \textit{yaw} axis by an angle $\alpha$ drawn from a uniform distribution $U(-\beta, + \beta)$ where $\beta \in \{\sfrac{\pi}{20}, \sfrac{\pi}{10}, \sfrac{\pi}{4}\}$. Simultaneously, we rotate the annotation $a(x_c, y_c, z_c, w, l, h, \theta)$ such that the augmented annotation $a^*$ has the form $a(x_c, y_c, z_c, w, l, h, \scriptstyle(\theta + \alpha)\mod 2\pi\textstyle)$. \\

\vspace{-3mm}
\subsubsection{Local Scaling} \label{local_scaling}\hfill\\
Local scaling~(Fig.~\ref{fig:local_scaling}) means that we are scaling every point $p(x, y, z) \in P$ in every direction by a scalar $s$ drawn from a uniform distribution $U(1-t, 1 + t)$ where $t \in \{0.05, 0.1, 0.25\}$ such that an augmented point $p^*$ has the form $p(s \cdot x, s \cdot y, s \cdot z)$. Simultaneously, we scale every annotation $a(x_c, y_c, z_c, w, l, h, \theta) \in A$ such that an augmented annotation $a^*$ has the form $a(s \cdot x_c, s \cdot y_c, s \cdot z_c, s \cdot w, s \cdot l, s \cdot h, \theta)$.

\subsection{GT-Filtering} \label{filter_augmentations}  
GT-Filtering is straightforward. One way to filter ground truth objects is to filter them based on their difficulty. In KITTI~\cite{KITTI} (and STF~\cite{STF} accordingly), there are three predefined difficulties: \textit{easy}, \textit{moderate} and \textit{hard}, where some annotations are also labeled as \textit{unknown}.

Another simple filter operation is to exclude annotations from training when they contain less than a certain amount of LiDAR points inside of them. In our experiments, we investigated a minimum threshold of 1, 5 and 10 points.

\begin{table*}

\resizebox{\textwidth}{!}{\begin{tabular}{ccccccccccccccccccrrr}
policy \#
& \rotatebox[origin=l]{90}{\textit{Global Translation}}
&  \rotatebox[origin=l]{90}{\textit{Global Rotation}}
&  \rotatebox[origin=l]{90}{\textit{Global Scaling}}
&  \rotatebox[origin=l]{90}{\textit{Random Flip}}   
&  \rotatebox[origin=l]{90}{\textit{Ground Removal}} 
&  \rotatebox[origin=l]{90}{} 
&  \rotatebox[origin=l]{90}{\textit{Local Translation}}   
&  \rotatebox[origin=l]{90}{\textit{Local Rotation}}   
&  \rotatebox[origin=l]{90}{\textit{Local Scaling}}   
&  \rotatebox[origin=l]{90}{} 
&  \rotatebox[origin=l]{90}{\textit{filter out unknown}} 
&  \rotatebox[origin=l]{90}{\textit{filter out hard}} 
&  \rotatebox[origin=l]{90}{\textit{filter out moderate}} 
&  \rotatebox[origin=l]{90}{\textit{filter by min. points}}   
&  \rotatebox[origin=l]{90}{} 
&  \rotatebox[origin=l]{90}{\textit{GT-Sampling}}  
&  \rotatebox[origin=l]{90}{}    
&  \textit{easy}
&  \shortstack[c]{\textbf{\textit{car}} \vspace{5pt} \\ $\textbf{3D mAP}_{\textbf{40}}$ \vspace{15pt} \\ \textit{moderate}}    
&  \textit{hard} \\

\hline \noalign{\vskip 1mm}

0 & \xmark & \xmark & \xmark & \xmark & \xmark 
&  & 
\xmark & \xmark & \xmark 
&  &
\xmark & \xmark & \xmark & \xmark 
&  & 
\xmark 
&  &  
71.98 &  59.29 & 55.94 \\ \noalign{\vskip 2mm}

1 & 0.1m & \xmark & \xmark & \xmark & \xmark 
&  & 
\xmark & \xmark & \xmark 
&  &
\xmark & \xmark & \xmark & \xmark 
&  & 
\xmark 
&  &  
$+$6.31 &  $+$5.55 & $+$3.93 \\ 

2 & \textbf{0.2m} & \xmark & \xmark & \xmark & \xmark 
&  & 
\xmark & \xmark & \xmark 
&  &
\xmark & \xmark & \xmark & \xmark 
&  & 
\xmark 
&  &  
\textbf{$+$7.21} &  \textbf{$+$6.64} & \textbf{$+$5.62} \\ 

3 & 0.4m & \xmark & \xmark & \xmark & \xmark 
&  & 
\xmark & \xmark & \xmark 
&  &
\xmark & \xmark & \xmark & \xmark 
&  & 
\xmark 
&  &  
$+$4.54 &  $+$5.18 & $+$4.47 \\ \noalign{\vskip 2mm}

4 & \xmark & $\pi/8$ & \xmark & \xmark & \xmark 
&  & 
\xmark & \xmark & \xmark 
&  &
\xmark & \xmark & \xmark & \xmark 
&  & 
\xmark 
&  &  
$+$7.97 &  $+$9.10 & $+$7.90 \\

5 & \xmark & $\pi/4$ & \xmark & \xmark & \xmark 
&  & 
\xmark & \xmark & \xmark 
&  &
\xmark & \xmark & \xmark & \xmark 
&  & 
\xmark 
&  &  
$+$10.34 &  $+$9.89 & $+$8.75 \\

6 & \xmark & $\boldsymbol{\pi/}\textbf{2}$ & \xmark & \xmark & \xmark 
&  & 
\xmark & \xmark & \xmark 
&  &
\xmark & \xmark & \xmark & \xmark 
&  & 
\xmark 
&  &  
\textbf{$+$11.51} &  \textbf{$+$10.73} & \textbf{$+$10.90} \\ \noalign{\vskip 2mm}

7 & \xmark & \xmark & \textbf{[0.95, 1.05]} & \xmark & \xmark 
&  & 
\xmark & \xmark & \xmark 
&  &
\xmark & \xmark & \xmark & \xmark 
&  & 
\xmark 
&  &  
$+$7.24 &  \textbf{$+$6.96} & \textbf{$+$5.43} \\

8 & \xmark & \xmark & [0.90, 1.10] & \xmark & \xmark 
&  & 
\xmark & \xmark & \xmark 
&  &
\xmark & \xmark & \xmark & \xmark 
&  & 
\xmark 
&  &  
$+$6.43 &  $+$5.72 & $+$4.87 \\

9 & \xmark & \xmark & [0.75, 1.25] & \xmark & \xmark 
&  & 
\xmark & \xmark & \xmark 
&  &
\xmark & \xmark & \xmark & \xmark 
&  & 
\xmark 
&  &  
\textbf{$+$8.10} &  $+$5.43 & $+$3.19 \\ \noalign{\vskip 2mm}

10 & \xmark & \xmark & \xmark & \color{green}\cmark & \xmark 
&  & 
\xmark & \xmark & \xmark 
&  &
\xmark & \xmark & \xmark & \xmark 
&  & 
\xmark 
&  &  
\textbf{$+$7.02} &  \textbf{$+$6.93} & \textbf{$+$7.58} \\ \noalign{\vskip 2mm}

11 & \xmark & \xmark & \xmark & \xmark & 1\% 
&  & 
\xmark & \xmark & \xmark 
&  &
\xmark & \xmark & \xmark & \xmark 
&  & 
\xmark 
&  &  
$+$0.48 & $+$1.79 & $+$1.17 \\

12 & \xmark & \xmark & \xmark & \xmark & 5\% 
&  & 
\xmark & \xmark & \xmark 
&  &
\xmark & \xmark & \xmark & \xmark 
&  & 
\xmark 
&  &  
\color{red}$-$2.75 &  \color{red}$-$1.51 & \color{red}$-$1.80 \\

13 & \xmark & \xmark & \xmark & \xmark & 10\% 
&  & 
\xmark & \xmark & \xmark 
&  &
\xmark & \xmark & \xmark & \xmark 
&  & 
\xmark 
&  &  
\color{red}$-$3.19 &  \color{red}$-$1.83 & \color{red}$-$1.97 \\

14 & \xmark & \xmark & \xmark & \xmark & 15\% 
&  & 
\xmark & \xmark & \xmark 
&  &
\xmark & \xmark & \xmark & \xmark 
&  & 
\xmark 
&  &  
\color{red}$-$4.72 &  \color{red}$-$3.83 & \color{red}$-$4.01 \\ \noalign{\vskip 2mm}

15 & \xmark & \xmark & \xmark & \xmark & \xmark 
&  & 
\textbf{0.05m} & \xmark & \xmark 
&  &
\xmark & \xmark & \xmark & \xmark 
&  & 
\xmark 
&  &  
$+$1.79 &  $+$0.71 & $+$1.14 \\

16 & \xmark & \xmark & \xmark & \xmark & \xmark 
&  & 
0.25m & \xmark & \xmark 
&  &
\xmark & \xmark & \xmark & \xmark 
&  & 
\xmark 
&  &  
$+$0.76 &  $+$0.52 & \color{red}$-$0.24 \\

17 & \xmark & \xmark & \xmark & \xmark & \xmark 
&  & 
0.50m & \xmark & \xmark 
&  &
\xmark & \xmark & \xmark & \xmark 
&  & 
\xmark 
&  &  
\color{red}$-$0.50 &  \color{red}$-$0.08 & \color{red}$-$1.35 \\

18 & \xmark & \xmark & \xmark & \xmark & \xmark 
&  & 
1.00m & \xmark & \xmark 
&  &
\xmark & \xmark & \xmark & \xmark 
&  & 
\xmark 
&  &  
\color{red}$-$0.77 &  \color{red}$-$0.33 & \color{red}$-$2.02 \\ \noalign{\vskip 2mm}

19 & \xmark & \xmark & \xmark & \xmark & \xmark 
&  & 
\xmark & $\boldsymbol{\pi/}\textbf{20}$ & \xmark 
&  &
\xmark & \xmark & \xmark & \xmark 
&  & 
\xmark 
&  &  
$+$6.08 &  \textbf{$+$4.66} & \textbf{$+$4.09} \\

20 & \xmark & \xmark & \xmark & \xmark & \xmark 
&  & 
\xmark & $\pi/10$ & \xmark 
&  &
\xmark & \xmark & \xmark & \xmark 
&  & 
\xmark 
&  &  
$+$4.77 &  $+$3.63 & $+$2.73 \\

21 & \xmark & \xmark & \xmark & \xmark & \xmark 
&  & 
\xmark & $\pi/4$ & \xmark 
&  &
\xmark & \xmark & \xmark & \xmark 
&  & 
\xmark 
&  &  
\textbf{$+$7.87} &  $+$4.57 & $+$3.06 \\ \noalign{\vskip 2mm}

22 & \xmark & \xmark & \xmark & \xmark & \xmark  
&  & 
\xmark & \xmark & \textbf{[0.95, 1.05]}
&  &
\xmark & \xmark & \xmark & \xmark 
&  & 
\xmark 
&  &  
\textbf{$+$4.98} &  \textbf{$+$3.99} & \textbf{$+$3.18} \\

23 & \xmark & \xmark & \xmark & \xmark & \xmark  
&  & 
\xmark & \xmark & [0.90, 1.10] 
&  &
\xmark & \xmark & \xmark & \xmark 
&  & 
\xmark 
&  &  
$+$4.82 &  $+$2.83 & $+$1.67 \\

24 & \xmark & \xmark & \xmark & \xmark & \xmark 
&  & 
\xmark & \xmark & [0.75, 1.25] 
&  &
\xmark & \xmark & \xmark & \xmark 
&  & 
\xmark 
&  &  
$+$4.79 &  $+$1.09 & \color{red}$-$1.49 \\ \noalign{\vskip 2mm}

25 & \xmark & \xmark & \xmark & \xmark & \xmark 
&  & 
\xmark & \xmark & \xmark
&  &
\color{green}\cmark & \xmark & \xmark & \xmark 
&  & 
\xmark 
&  &  
\color{red}$-$0.16 &  $+$0.67 & $+$0.54 \\

26 & \xmark & \xmark & \xmark & \xmark & \xmark 
&  & 
\xmark & \xmark & \xmark
&  &
\color{green}\cmark & \color{green}\cmark & \xmark & \xmark 
&  & 
\xmark 
&  &  
$+$1.19 &  $+$0.09 & $+$1.16 \\

27 & \xmark & \xmark & \xmark & \xmark & \xmark 
&  & 
\xmark & \xmark & \xmark
&  &
\color{green}\cmark & \color{green}\cmark & \color{green}\cmark & \xmark 
&  & 
\xmark 
&  &  
$+$1.79 &  $+$0.14 & \color{red}$-$0.45 \\ \noalign{\vskip 2mm}

28 & \xmark & \xmark & \xmark & \xmark & \xmark 
&  & 
\xmark & \xmark & \xmark
&  &
\xmark & \xmark & \xmark & \textbf{1}
&  & 
\xmark 
&  &  
$+$0.96 &  $+$1.01 & $+$0.67 \\

29 & \xmark & \xmark & \xmark & \xmark & \xmark 
&  & 
\xmark & \xmark & \xmark
&  &
\xmark & \xmark & \xmark & 5 
&  & 
\xmark 
&  &  
\color{red}$-$0.10 &  $+$0.29 & $+$0.84 \\

30 & \xmark & \xmark & \xmark & \xmark & \xmark 
&  & 
\xmark & \xmark & \xmark
&  &
\xmark & \xmark & \xmark & 10 
&  & 
\xmark 
&  &  
$+$0.76 &  $+$0.52 & \color{red}$-$0.48 \\ \noalign{\vskip 2mm}

31 & \xmark & \xmark & \xmark & \xmark & \xmark 
&  & 
\xmark & \xmark & \xmark
&  &
\xmark & \xmark & \xmark & \xmark 
&  & 
\textbf{5}
&  &  
$+$2.71 &  $+$2.86 & $+$2.83 \\

32 & \xmark & \xmark & \xmark & \xmark & \xmark 
&  & 
\xmark & \xmark & \xmark
&  &
\xmark & \xmark & \xmark & \xmark 
&  & 
10 
&  &  
$+$1.29 &  $+$2.59 & $+$1.70 \\

33 & \xmark & \xmark & \xmark & \xmark & \xmark 
&  & 
\xmark & \xmark & \xmark
&  &
\xmark & \xmark & \xmark & \xmark 
&  & 
15 
&  &  
\color{red}$-$0.19 &  $+$1.29 & $+$0.64 \\ 

34 & \xmark & \xmark & \xmark & \xmark & \xmark 
&  & 
\xmark & \xmark & \xmark
&  &
\xmark & \xmark & \xmark & \xmark 
&  & 
20 
&  &  
$+$1.72 &  $+$1.31 & $+$0.48 \\

35 & \xmark & \xmark & \xmark & \xmark & \xmark 
&  & 
\xmark & \xmark & \xmark
&  &
\xmark & \xmark & \xmark & \xmark 
&  & 
25 
&  &  
$+$0.51 &  $+$1.01 & $+$0.32 \\ \noalign{\vskip 2mm}

\color{magenta}36 & \color{magenta}\textbf{0.2m} & \color{magenta}$\pi/4$ & \color{magenta}\textbf{[0.95, 1.05]} & \color{magenta}\cmark & \color{magenta}\xmark 
&  & 
\color{magenta}0.25m & \color{magenta}$\boldsymbol{\pi/}\textbf{20}$ & \color{magenta}\xmark
&  &
\color{magenta}\cmark & \color{magenta}\xmark & \color{magenta}\xmark & \color{magenta}5 
&  & 
\color{magenta}15
&  &  
\color{magenta}$+$15.75 &  \color{magenta}$+$17.72 & \color{magenta}$+$16.59 \\ \noalign{\vskip 2mm}

37 & \textbf{0.2m} & $\pi/4$ & \textbf{[0.95, 1.05]} & \cmark & \xmark 
&  & 
0.25m & $\boldsymbol{\pi/}\textbf{20}$ & \xmark
&  &
\cmark & \xmark & \xmark & 5 
&  & 
\xmark
&  &  
$+$11.45 &  $+$11.06 & $+$10.60 \\ \noalign{\vskip 2mm}

38 & \xmark & $\boldsymbol{\pi/}\textbf{2}$ & \xmark & \xmark & \xmark 
&  & 
\xmark & \xmark & \xmark
&  &
\xmark & \xmark & \xmark & \xmark 
&  & 
15
&  &  
$+$13.14 &  $+$14.88 & $+$13.88 \\ \noalign{\vskip 2mm}

\color{cyan}39 & \color{cyan}\textbf{0.2m} & \color{cyan}$\pi/4$ & \color{cyan}\textbf{[0.95, 1.05]} & \color{cyan}\cmark & \color{cyan}\xmark 
&  & 
\color{cyan}\xmark & \color{cyan}$\boldsymbol{\pi/}\textbf{20}$ & \color{cyan}\xmark 
&  &
\color{cyan}\cmark & \color{cyan}\xmark & \color{cyan}\xmark & \color{cyan}5 
&  & 
\color{cyan}15
&  &  
\color{cyan}$+$15.95 &  \color{cyan}$+$18.05 & \color{cyan}$+$18.30 \\

\color{cyan}40 & \color{cyan}\textbf{0.2m} & \color{cyan}$\pi/4$ & \color{cyan}\textbf{[0.95, 1.05]} & \color{cyan}\cmark & \color{cyan}\xmark 
&  & 
\color{cyan}\xmark & \color{cyan}$\boldsymbol{\pi/}\textbf{20}$ & \color{cyan}\textbf{[0.95, 1.05]}
&  &
\color{cyan}\cmark & \color{cyan}\xmark & \color{cyan}\xmark & \color{cyan}5 
&  & 
\color{cyan}15
&  &  
\color{cyan}$+$16.08 &  \color{cyan}$+$18.53 & \color{cyan}$+$\underline{18.51} \\

\color{cyan}41 & \color{cyan}\textbf{0.2m} & \color{cyan}$\pi/4$ & \color{cyan}\textbf{[0.95, 1.05]} & \color{cyan}\cmark & \color{cyan}\xmark 
&  & 
\color{cyan}\xmark & \color{cyan}$\boldsymbol{\pi/}\textbf{20}$ & \color{cyan}\textbf{[0.95, 1.05]}
&  &
\color{cyan}\cmark & \color{cyan}\cmark & \color{cyan}\xmark & \color{cyan}5 
&  & 
\color{cyan}15
&  &  
\color{cyan}\textbf{$+$16.95} &  \color{cyan}\textbf{$+$19.20} & \color{cyan}\textbf{$+$18.54} \\

\color{cyan}42 & \color{cyan}\textbf{0.2m} & \color{cyan}$\boldsymbol{\pi/}\textbf{2}$ & \color{cyan}\textbf{[0.95, 1.05]} & \color{cyan}\cmark & \color{cyan}\xmark 
&  & 
\color{cyan}\xmark & \color{cyan}$\boldsymbol{\pi/}\textbf{20}$ & \color{cyan}\textbf{[0.95, 1.05]}
&  &
\color{cyan}\cmark & \color{cyan}\cmark & \color{cyan}\xmark & \color{cyan}5 
&  & 
\color{cyan}15
&  &  
\color{cyan}$+$\underline{16.52} &  \color{cyan}$+$\underline{18.83} & \color{cyan}$+$17.77 \\

\end{tabular}}
\caption{Results of our extensive augmentation study on the KITTI~\cite{KITTI} validation set. Most~significant~improvements~in~\textbf{bold}. Augmentation policy of PointPillars~\cite{PointPillars} in \color{magenta}magenta (\#36)\color{black}. Our improved augmentation policies in \color{cyan}cyan (\#39$-$42)\color{black}.}
\label{table:results}
\end{table*}

\begin{table}[b]
\centering
\begin{tabular}{lcccc}
method                       & policy \#   & \multicolumn{1}{c}{\textit{easy}}  & \multicolumn{1}{c}{\textit{moderate}} & \multicolumn{1}{c}{\textit{hard}}  \\ \hline \noalign{\vskip 1mm}
\multirow{2}{*}{VoxelRCNN~\cite{VRCNN}}   & 36          & 91.89                              & 84.36                                 & 82.13                              \\
                             & 41          & \textbf{92.62}                     & \textbf{85.14}                        & \multicolumn{1}{c}{\textbf{82.31}} \\ \noalign{\vskip 1mm}
\multirow{2}{*}{PV-RCNN~\cite{PV-RCNN}}     & 36          & 91.28                              & 84.06                                 & 81.70                              \\
                             & 41          & \textbf{92.55}                     & \textbf{85.02}                        & \textbf{82.00}                     \\ \noalign{\vskip 1mm}
\multirow{2}{*}{CenterPoint~\cite{CP}} & 36          & \multicolumn{1}{c}{91.63}          & 83.92                                 & \textbf{81.67}                     \\
                             & 41          & \multicolumn{1}{c}{\textbf{92.04}} & \textbf{84.69}                        & \multicolumn{1}{c}{81.62}          \\ \noalign{\vskip 1mm}
\multirow{2}{*}{Part-A²~\cite{PartA2}}     & 36          & \multicolumn{1}{c}{91.83}          & 82.49                                 & 80.17                              \\
                             & 41          & \multicolumn{1}{c}{\textbf{92.38}} & \textbf{83.24}                        & \textbf{80.28}                     \\ \noalign{\vskip 1mm}
\multirow{2}{*}{SECOND~\cite{SECOND}}      & 36          & 89.71                              & 80.77                                 & \multicolumn{1}{l}{77.75}          \\
                             & 41          & \textbf{90.11}                     & \textbf{81.23}                        & \multicolumn{1}{l}{\textbf{76.36}}
\end{tabular}
\caption{Further 3D mAP$_{40}$ results of the \textit{car} class on the KITTI~\cite{KITTI} validation set.}
\label{tab:kitti_results}
\end{table}

\begin{table}[b]
\centering
\begin{tabular}{lcccc}
method                       & policy \# & \textit{easy}                      & \multicolumn{1}{c}{\textit{moderate}} & \textit{hard}                      \\ \hline \noalign{\vskip 1mm}
\multirow{2}{*}{VoxelRCNN~\cite{VRCNN}}   & 36        & \multicolumn{1}{r}{39.86}          & 39.39                                 & \multicolumn{1}{r}{36.20}          \\
                             & 41        & \multicolumn{1}{r}{\textbf{40.77}} & \textbf{41.12}                        & \multicolumn{1}{r}{\textbf{36.55}} \\ \noalign{\vskip 1mm}
\multirow{2}{*}{PV-RCNN~\cite{PV-RCNN}}     & 36        & \textbf{41.66}                     & 41.94                                 & 37.60                              \\
                             & 41        & \multicolumn{1}{r}{41.23}          & \textbf{42.67}                        & \textbf{38.41}                     \\ \noalign{\vskip 1mm}
\multirow{2}{*}{CenterPoint~\cite{CP}} & 36        & 41.80                              & 42.45                                 & \textbf{38.42}                     \\
                             & 41        & \textbf{43.02}                     & \textbf{42.87}                        & 38.30                              \\ \noalign{\vskip 1mm}
\multirow{2}{*}{Part-A²~\cite{PartA2}}     & 36        & 39.84                              & 39.74                                 & \multicolumn{1}{r}{36.06}          \\
                             & 41        & \textbf{41.01}                     & \textbf{40.68}                        & \multicolumn{1}{r}{\textbf{37.52}} \\ \noalign{\vskip 1mm}
\multirow{2}{*}{SECOND~\cite{SECOND}}      & 36        & \multicolumn{1}{r}{\textbf{38.32}} & 38.16                                 & \multicolumn{1}{l}{35.17}          \\
                             & 41        & \multicolumn{1}{r}{37.27}          & \textbf{38.84}                        & \multicolumn{1}{l}{\textbf{35.84}}
\end{tabular}
\caption{3D mAP$_{40}$ results of the \textit{car} class on the challenging STF~\cite{STF} validation set.}
\label{tab:dense_results}
\end{table}

\subsection{GT-Sampling} \label{GT-Sampling}  
GT-Sampling means that we are trying to sample additional ground truth objects, \eg cars, from other scenes into the current scene. In order to do so, one has to iterate over all annotations once and construct a database of annotations and their corresponding points. This ``trick" also aims at balancing the ``foreground" vs. ``background" imbalance described earlier in Sec.~\ref{ground_removal} . During training, one only has to check whether the additionally sampled annotations from the database do not collide with any of the ones originally present in the current scene. In our experiments we tried to additionally sample 5, 10, 15, 20 and 25 annotations. If there is a collision, those annotations are discarded and only the ones that do not collide are kept. This means for example in the setting where we try to sample up to 15 cars, at most 15 annotations are additionally sampled.

\vspace{5mm}
\section{Experiments} \label{experiments}
In this Section, we present the findings of our extensive augmentation study summarized in Table~\ref{table:results},~\ref{tab:kitti_results},~\ref{tab:dense_results}, and~\ref{tab:comparison}.

We always report 3D mAP$_{40}$ proposed by~\cite{Simonelli_2019_ICCV}, the mean average precision computed over 40 instead of 11 recall operation points originally proposed by the Pascal~VOC~benchmark~\cite{PascalVOC}. 
For more details, please refer to~\cite{Simonelli_2019_ICCV}. 
In this paper we focus on the \textit{car} class and carry out our main experiments on the KITTI~\cite{KITTI} validation set, following the common practice proposed in \cite{split} to split the official training set into 3,712 training and 3,769 validation samples. 

In order to investigate all the potential augmentation techniques introduced in Section~\ref{method}, we have to establish a proper baseline where no augmentation technique is applied (policy~\#0). To reduce the impact of randomness and to have a more meaningful comparison, we carried out all experiments listed in Table~\ref{table:results},~\ref{tab:kitti_results},~\ref{tab:dense_results}, and~\ref{tab:comparison} exactly three times and report the numbers of the training run which has the highest 3D mAP$_{40}$ score on the \textit{moderate} difficulty, the same metric that is used for ranking submissions on the \href{http://www.cvlibs.net/datasets/kitti/eval\_object.php?obj\_benchmark=3d}{official KITTI leader-board}. 

Diving into Table~\ref{table:results}, our first finding is, that there is a huge gap (17.72\%) between the baseline with no augmentation (policy~\#0) and the augmentation policy used in PointPillars~\cite{PointPillars} (policy~\#36). This means that almost $\sfrac{1}{4}$ (23\%) of the performance of PointPillars~\cite{PointPillars} can be attributed to its sophisticated augmentation policy.

If we look at individual augmentation techniques, most notably we see that \textit{Global Rotation} (policy~\#4-6) is the most effective augmentation with an increase of up and above 10\% (policy~\#6). Followed by three other \textit{global} augmentation techniques, \textit{Global Translation} (policy~\#1-3), \textit{Global Scaling} (policy~\#7-9) and \textit{Random Flip} (policy~\#10), all contributing with a performance boost of around 7\%. 

To show how effective \textit{Global Rotation} is as an augmentation technique, we can compare our best performing \textit{Global Rotation} policy (policy~\#6), with the advanced augmentation policy of PointPillars~\cite{PointPillars} where only \textit{GT-Sampling} is removed (resulting in policy~\#37), and we can see that the two policies perform similarly.  

Only adding \textit{GT-Sampling} to policy \#6 (resulting in policy \#38) we can already half the gap to the performance of the sophisticated policy (two vs. nine augmentation techniques) utilized in PointPillars~\cite{PointPillars} (policy \#36). On the other hand, \textit{GT-Sampling} applied by itself (policy \#31-35) does not help as much as other individual augmentation techniques. It only shines in combination with other augmentation techniques, showcasing that the combination of the different augmentation techniques are not simply additive.

Investigating the remaining \textit{global} augmentation technique \textit{Ground Removal} (policy~\#11-14), which is intuitive and often performed in classical LiDAR pipelines~\cite{Kabzan_AMZ}, is rather disadvantageous than beneficial in our experiments. We speculate that this could be due to two things. First, by removing the lowest points of the scene (in such a naive way) a lot of potentially crucial neighborhood context around ``foreground" objects gets deleted. Second, it could also mean that ``background" points quite generally encode more valuable information than what is the established view on that matter. In order to clarify this though, further experiments would be necessary (see future work in Section~\ref{conclusion}).

In general we see that \textit{local} augmentation techniques are not as effective as \textit{global} augmentation techniques. Although, while \textit{Local Rotation} (policy~\#19-21) and \textit{Local Scaling} (policy~\#22-24) can be considered beneficial, \textit{Local Translation} (policy~\#15-18) contrarily is clearly not as beneficial as the other two \textit{local} augmentation techniques and can even hurt performance if applied too aggressively. 

\textit{GT-Filtering} (policy~\#25-30) also does not make a big difference, interesting is maybe only the results of policy~\#26, where we filter all \textit{hard} examples during training and still get a higher performance on them during evaluation (+1.16\%). This could mean that if we keep \textit{hard} examples during training, the network might get too distracted by those seemingly diverse \textit{hard} examples and is not able extract distinct features from them, thereby hurting its generalisation capabilities.

If we now take all those insights from policy~\#1-35 and apply them to the policy of PointPillars~\cite{PointPillars} (policy~\#36) we can further boost its performance. First, by removing the potentially hurtful \textit{Local Translation} (resulting in policy~\#39), then adding \textit{Local Scaling} (resulting in policy~\#40) and finally also leaving out \textit{hard} examples during training (resulting in policy~\#41), we can further improve PointPillars'~\cite{PointPillars} results by up to 1.5\% on \textit{moderate} and almost 2\% on \textit{hard} cars.
Lastly in policy~\#42, where we combine all the individual best augmentation techniques, does not yield the best overall result, showcasing (again) that the combination of individual techniques are not just simply additive.

Table~\ref{tab:kitti_results} shows that our findings from Table~\ref{table:results} are not just limited to PointPillars~\cite{PointPillars}, but generalize well to five other state of the art 3D Object Detection methods. 

Table~\ref{tab:dense_results} demonstrates that the findings from Table~\ref{table:results} are also not just limited to the KITTI~\cite{KITTI} dataset, but transfer well to the challenging STF~\cite{STF} dataset. We believe that this is also applicable to other LiDAR datasets, such as nuScenes~\cite{nuScenes} and the Lyft~Level5~Perception~Dataset~\cite{LyftL5Dataset}. Even though these datasets are captured with different LiDAR sensors and hence can have drastically different number of overall LiDAR points per scene (Lyft~Level5~Perception~Dataset~\cite{LyftL5Dataset} with up to 190,000 points per scene vs. nuScenes~\cite{nuScenes} with only around 35,000 points per scene on average), the main issue, the imbalance between ``foreground" and ``background" points remains the same for all current LiDAR datasets. 

Table~\ref{tab:comparison} shows our best augmentation policy compared to related work. We can see that our policy \#41 can not be outperformed by~\cite{Part-Aware},~\cite{zheng2021se}, or~\cite{Pattern-Aware} applied to policy \#36, the common default strategy introduced in PointPillars~\cite{PointPillars}. Only applying~\cite{Part-Aware},~\cite{zheng2021se}, or~\cite{Pattern-Aware} on top of our policy \#41 in some cases can marginally push the performance even higher.





\begin{table}[t]
\centering
\begin{tabular}{llccc}
method                       & \multicolumn{1}{c}{policy \#} & \multicolumn{1}{c}{\textit{easy}}  & \multicolumn{1}{c}{\textit{moderate}} & \multicolumn{1}{c}{\textit{hard}} \\ \hline \noalign{\vskip 1mm}
\multirow{7}{*}{VoxelRCNN~\cite{VRCNN}}   & 36 +~\cite{Part-Aware}                       & 91.93                              & 84.04                                 & 81.95                             \\
                             & 36 +~\cite{zheng2021se}                       & 91.73                              & 84.05                                 & 81.56                             \\
                             & 36 +~\cite{Pattern-Aware}                       & 91.52                              & 84.26                                 & 82.05                             \\
                             & 41                            & \multicolumn{1}{r}{\textbf{92.62}} & {\ul 85.14}                           & \multicolumn{1}{c}{{\ul 82.31}}   \\
                             & 41 +~\cite{Part-Aware}                       & {\ul 92.55}                        & \textbf{85.46}                        & \textbf{82.65}                    \\
                             & 41 +~\cite{zheng2021se}                       & 92.52                              & 84.90                                 & 82.16                             \\
                             & 41 +~\cite{Pattern-Aware}                       & 92.46                              & 84.99                                 & 82.00                             \\ \noalign{\vskip 3mm}
\multirow{7}{*}{PV-RCNN~\cite{PV-RCNN}}     & 36 +~\cite{Part-Aware}                       & 90.60                              & 84.06                                 & 81.98                             \\
                             & 36 +~\cite{zheng2021se}                       & 91.78                              & 84.40                                 & 82.16                             \\
                             & 36 +~\cite{Pattern-Aware}                       & 91.80                              & 84.34                                 & 82.12                             \\
                             & 41                            & \multicolumn{1}{r}{\textbf{92.55}} & {\ul 85.02}                           & \multicolumn{1}{r}{82.00}         \\
                             & 41 +~\cite{Part-Aware}                       & 92.25                              & 84.91                                 & 81.70                             \\
                             & 41 +~\cite{zheng2021se}                       & 92.06                              & \textbf{85.14}                        & \textbf{82.41}                    \\
                             & 41 +~\cite{Pattern-Aware}                       & {\ul 92.27}                        & 84.97                                 & {\ul 82.29}                       \\

\end{tabular}
\caption{Comparison to the augmentation methods introduced in ~\cite{Part-Aware},~\cite{zheng2021se}, and~\cite{Pattern-Aware}. Again, we present 3D mAP$_{40}$ of the \textit{car} class on the KITTI~\cite{KITTI} validation set.}
\label{tab:comparison}
\end{table}

\balance

\section{Conclusion \& Future Work} \label{conclusion}
In this work, we provide many insights into the effectiveness of different augmentation techniques for LiDAR based 3D Object Detection. 
We demonstrate that our improved augmentation policy \#41 transfers well to other 3D Object Detection methods and are not just limited to the KITTI~\cite{KITTI} dataset. We hope that other practitioners can now take our findings off the shelf and apply them successfully to their work. Further, we uncovered how important data augmentation for LiDAR based 3D Object Detection really is and that it seemingly plays a significant role for many state of the art 3D Object Detection methods.

In future work, we want to investigate why \textit{Ground Removal} can harm 3D Object Detection performance. One explanation for the deterioration could be that significant scene / neighborhood context is getting destroyed by removing the points belonging to the ground. However, more experiments are needed to test this hypothesis, 
\eg by estimating the ground plane~\cite{paigwar:hal-02927350} or using semantic ground truth information provided by SemanticKITTI~\cite{SemanticKITTI} or nuScenes-lidarseg~\cite{fong2021panoptic}.

\bibliographystyle{IEEEtran}
\bibliography{root}

\end{document}